\documentclass[letterpaper, 10 pt, conference]{ieeeconf}  

\IEEEoverridecommandlockouts                              

\usepackage{graphicx} 
\usepackage[ruled,vlined,linesnumbered]{algorithm2e}
\usepackage{subcaption}

\newtheorem{theorem}{Theorem}

\title{\LARGE \bf
Optimal and Bounded Suboptimal Any-Angle Multi-agent Pathfinding
}

\author{Konstantin Yakovlev$^{1}$, Anton Andreychuk$^{2}$, Roni Stern$^{3}$
\thanks{\emph{This is a pre-print version of the paper accepted to IROS 2024. Its main body is similar to the camera-ready version of the conference paper. In addition this pre-print contains Appendix.}}%
\thanks{$^{1}$Konstantin Yakovlev is with FRC CSC RAS, AIRI, HSE University, MIPT 
        {\tt\small yakovlev@isa.ru}}%
\thanks{$^{2}$Anton Andreychuk is with AIRI
        {\tt\small andreychuk@airi.net}}%
\thanks{$^{3}$Roni Stern is with Ben-Gurion University of the Negev
        {\tt\small sternron@bgu.ac.il}}%
}

\begin{document}

\maketitle
\thispagestyle{empty}
\pagestyle{empty}

\begin{abstract}

Multi-agent pathfinding (MAPF) is the problem of finding a set of conflict-free paths for a set of agents. 
Typically, the agents' moves are limited to a pre-defined graph of possible locations and allowed transitions between them, e.g. a 4-neighborhood grid.  
We explore how to solve MAPF problems when each agent can move between any pair of possible locations as long as traversing the line segment connecting them does not lead to a collision with the obstacles. 
This is known as any-angle pathfinding.  
We present the first optimal any-angle multi-agent pathfinding algorithm. 
Our planner is based on the Continuous Conflict-based Search (CCBS) algorithm and an optimal any-angle variant of the Safe Interval Path Planning (TO-AA-SIPP). 
The straightforward combination of those, however, scales poorly since any-angle path finding induces search trees with a very large branching factor. 
To mitigate this, we adapt two techniques from classical MAPF to the any-angle setting, namely Disjoint Splitting and Multi-Constraints. 
Experimental results on different combinations of these techniques 
show they enable solving over 30\% more problems than the vanilla combination of CCBS and TO-AA-SIPP. 
In addition, we present a bounded-suboptimal variant of our algorithm, that enables trading runtime for solution cost in a controlled manner. 

\end{abstract}

\section{INTRODUCTION}
Multi-agent pathfinding (MAPF) is the problem of finding a set of conflict-free paths for a set of agents. 
Its applications include automated warehouses~\cite{wurman2008coordinating}, traffic control~\cite{parks2022intersection}, digital entertainment~\cite{ma2017feasibility}, etc. Different optimization variants of MAPF have proven to be NP-Hard~\cite{yu2016optimal}, yet efficient optimal MAPF solvers have been proposed such as  Conflict Based Search (CBS)~\cite{sharon2015cbs}, SAT-MDD~\cite{surynek2016efficient}, BCP~\cite{lam2022bcp} and others.

\begin{figure}[t!]
    \centering
    \includegraphics[width=1\linewidth]{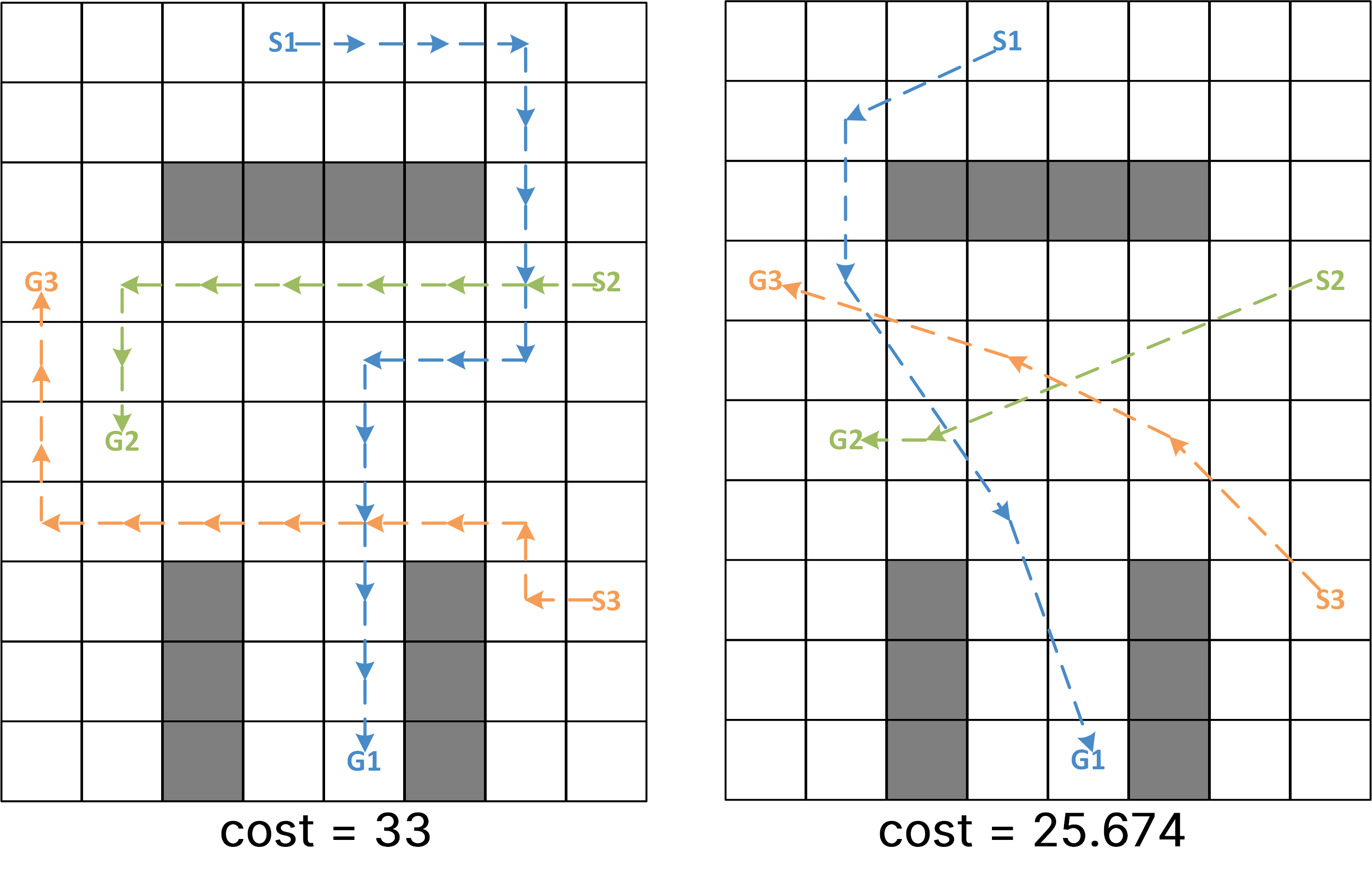}
    \caption{Two optimal solutions of the same MAPF instance: the one composed of the cardinal moves only (left) and the one with any-angle moves (right). The cost of the latter is $22\%$ lower.}
    \label{fig:vabstract}
\end{figure}

Most prior work focused on the classical version of the problem which makes many simplifying assumptions~\cite{stern2019multi}. These assumptions include \emph{(i)} the agents' moves are limited to a given sparse graph of possible locations and the transitions between them, and \emph{(ii)} all moves take one time unit. Indeed, these limitations are not well-suited for a variety of robotic applications and, thus, some works have begun to explore how to solve MAPF without them. In particular, the Continuous Conflict-based Search (CCBS) algorithm avoids the need to discretize time and can handle agent actions with different durations~\cite{andreychuk2022multi}. Still, CCBS as well as many other previously proposed optimal MAPF solvers rely on the first assumption. We focus on relaxing it, allowing each agent to move between \textit{any} pair of locations -- see Fig.~\ref{fig:vabstract}. This type of pathfinding is known as \emph{any-angle pathfinding}~\cite{nash2007}. 

Algorithms such as Anya~\cite{harabor2013optimal} and TO-AA-SIPP~\cite{yakovlev2021towards} have been proposed for optimal single-agent any-angle path finding, and a suboptimal any-angle MAPF solver was proposed~\cite{yakovlev2017any}. We propose AA-CCBS, which integrates CCBS and TO-AA-SIPP to form the first practical any-angle MAPF algorithm that is guaranteed to return cost-optimal solutions.

Unfortunately, AA-CCBS scales poorly since any-angle path finding induces search trees with a very large branching factor. 
To mitigate this, we propose several enhancements to AA-CCBS based on techniques from classical MAPF, namely disjoint splitting (DS)~\cite{li2019disjoint} and multi-constraints (MC)~\cite{walker2020generalized}. In particular, we suggest several novel variants of how MC can be applied for our any-angle setup, without compromising the theoretical guarantees.

We conduct a thorough empirical evaluation of different variants of AA-CCBS across standard benchmarks. 
The results show that AA-CCBS with our enhancements works very well, solving significantly more problems under a given time limit than vanilla AA-CCBS. 
We also show how AA-CCBS can be generalized to return solutions with bounded suboptimality, 
allowing a controlled way to trade off runtime for solution quality. 
Our experimental results demonstrate that allowing even a small amount of suboptimality allows solving many more problems. 

\section{RELATED WORKS}

Lifting certain classical MAPF assumptions has already been explored. 
Yakovlev and Andreychuk~\cite{yakovlev2017any} proposed an any-angle MAPF planner based on prioritized planning. This planner was later generalized to support robots of different sizes and moving speeds~\cite{yakovlev2019prioritized}. 
Other techniques were proposed to support kinematic constraints~\cite{ma2019lifelong, honig2016multi,ali2021prioritized}. 
Yan and Li~\cite{YanRAL24} proposed an involved three-stage solver based on Priority-Based Search~\cite{ma2019searching} that handles kinematic constraints as well as accelerating and decelerating actions. 
All these solvers, unlike ours, are not optimal.

Optimal MAPF solvers that go beyond classical MAPF setting are often adaptations of CBS~\cite{sharon2015cbs}. 
Solis et al.~\cite{solis2021representation} proposed to construct a roadmap for each robot and plan using CBS. 
A dedicated technique is suggested to decide when to re-construct these roadmaps. 
A combination of CBS with trajectory optimization methods was suggested to construct collision-free paths for a team of heterogeneous agents (quadrotors)~\cite{debord2018trajectory,honig2018trajectory}. 
Kinodynamic variants of CBS were considered in~\cite{kottinger2022conflict,moldagalieva2023db,wen2022cl}. 
In this work, we also leverage CBS to conduct multi-agent any-angle path finding with optimality guarantees.

\section{PROBLEM STATEMENT}
Consider a set of vertices $V$ residing in the metric space, e.g., a grid where the vertices are the set of centers of the grid cells. There are $n$ designated start and goal vertices, and $n$ agents initially located at the start vertices.
The agents are modeled as disks of a certain radius. Two types of actions for each agent are possible: wait at the current vertex (for an arbitrary amount of time) and move between the vertices. 
A move is valid iff the given line-of-sight function, $los: V \times V \rightarrow \{true, false\}$, returns $true$. This function checks whether an agent can follow a straight-line segment connecting the vertices without colliding with an obstacle, e.g., whether the set of grid cells swept by the disk-shaped agent contains only free cells. The cost of each action is its duration. The duration of a wait action may be arbitrary. The duration of a move action equals the distance between the vertices comprising this move (we assume that the agents start/stop instantaneously and move with the unit speed).

An individual plan for an agent is a sequence of timed actions $\pi=((a_0, t_0), (a_1, t_1), ..., (a_n, t_n))$, where $a_i$ is a move action, $t_i$ is the starting time of this action, and each action must start where the previous one ends. The wait actions are encoded implicitly in the plan, i.e. if $t_i + dur(a_i) < t_{i+1}$ then the agent is waiting at the endpoint of $a_i$ for a time of $t_{i+1} - t_i - dur(a_i)$. Here $dur(a_i)$ denotes the duration of action $a_i$.
The plans of distinct agents are said to be in conflict if there exists a time moment when the distance between the agents executing these plans is less than the sum of their radii. 
The any-angle multi-agent pathfinding problem (AA-MAPF) asks to find a set of $n$ plans transferring the agents from their start vertices to the goal ones, such that each pair of plans is collision-free. 
The cost of the individual plan, $c(\pi)$, is the time when the agent reaches its goal. 
We wish to solve the problem optimally w.r.t. the sum-of-cost objective, which is the sum of durations of the agents' plans, $SOC=\sum_{i=1}^n{c(\pi_i)}$. Adapting to other optimization criteria such as makespan is also possible. 

\section{BACKGROUND}
Our planner is based on a combination of CCBS,~\cite{andreychuk2022multi} and TO-AA-SIPP~\cite{yakovlev2021towards}. Next, we briefly introduce these methods.

\subsection{CCBS} 

CCBS is a variant of CBS~\cite{sharon2015cbs} adapted to non-discretized timeline. 
It reasons over the continuous time-intervals instead of distinct isolated time steps. 
CCBS works by finding plans for each agent separately, detecting \emph{conflicts} between these plans, and resolving them by replanning for the individual agents subject to specific \emph{constraints}. 
A conflict between the two agents in CCBS is defined by $(a_i, a_j, t_i, t_j)$, representing that a collision occurs when executing the actions $a_i$ and $a_j$ at time moments $t_i$ and $t_j$, respectively. 
CCBS resolves a conflict by imposing a constraint on an agent and replan for that agent. The constraint imposed by CCBS is a tuple $(a, [t, t'))$, stating that the agent is prohibited from taking action $a$ in the time range $[t, t')$. The latter is called the \emph{unsafe interval}. 
Several approaches were considered to compute $t'$, the first time moment an agent can safely start executing the action, including closed-loop formulas~\cite{walker2019collision}. 

To guarantee optimality, CCBS runs two search processes: 
a high-level search to choose which agent to constraint 
and a low-level search to find a path for that agent that satisfies its constraints. 
CCBS uses a best-first search for the high-level search and 
the Safe-Interval Path Planning (SIPP) algorithm~\cite{phillips2011sipp} for the low-level search. SIPP is a specific adaptation of A* for space-time domains. It was originally proposed for single-agent pathfinding among dynamic obstacles, but can be easily adapted to handle CCBS constraints, as the latter can be viewed as dynamic obstacles that temporarily appear in the workspace. 

\subsection{TO-AA-SIPP} 

TO-AA-SIPP is a variant of SIPP that aims at finding an optimal any-angle path for an agent navigating in an environment with dynamic obstacles. Its search node is identified by a tuple $(v, [t_l, t_u])$, where $v$ is a graph vertex and $[t_l, t_u]$ is a safe time interval that dictates when the agent may safely reside at $v$. 

TO-AA-SIPP (unlike SIPP) does not iteratively build a search tree by expanding the search nodes. Instead, it generates all the search nodes beforehand and iteratively tries to identify the correct parent relationships between the nodes to obtain an optimal solution. TO-AA-SIPP was proven to be complete and optimal.

\section{AA-CCBS}
To solve AA-MAPF, we propose to use TO-AA-SIPP as the low-level planner for CCBS. We call this algorithm AA-CCBS. 
Unfortunately, AA-CCBS struggles to solve even simple instances.  E.g., the small AA-MAPF instance with 3 agents depicted in Fig.~\ref{fig:vabstract} requires $6,660$ high-level iterations and consequently the same number of calls to TO-AA-SIPP. 
To enable AA-CCBS to scale more gracefully, we propose two enhancements: Multi-Constrains (MC) and Disjoint Splitting (DS).

\subsection{Multi-Constraints in Any-Angle MAPF}
CCBS resolves a conflict by adding a single constraint to one of the agents and replanning. 
However, adding multiple constraints (multi-constraints) when resolving a single conflict can reduce the number of high-level search iterations~\cite{li2019multi,walker2020generalized}. In particular, Walker et al.~\cite{walker2020generalized} proposed the Time-Annotated Biclique (TAB) method for adding multiple constraints in domains with non-uniform actions.  

For two conflicting actions, $(a_i,t_i)$ and $(a_j, t_j)$, TAB iterates over all actions that have the same source vertices as either $a_i$ or $a_j$ and identifies the subsets of them, $A_i$ and $A_j$, that are \emph{mutually conflicting}, i.e. each pair of actions from $A_i \times A_j$ lead to a conflict (if they start at $t_i$, $t_j$ respectively). 
The multi-constraint (MC) added to the first agent comprises the constraints $(a_i, [t, t'])$ where $[t, t']$ is the largest time interval that is \textit{fully included} into the unsafe intervals induced by $a_i$ and \textit{all} $a_j \in A_j$. The MC added to the second agent is defined similarly. 
TAB is applicable in AA-CCBS. We call this variant MC1.

\begin{figure}[t]
    \centering
    \includegraphics[width=0.85\linewidth]{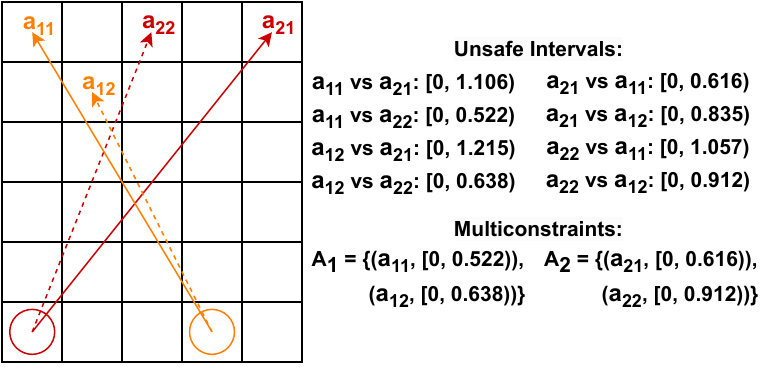}
    \caption{An example of vanilla AA-CCBS multi-constraint.}
    \label{fig:MC-example}
\end{figure}

\vspace{3pt}

\noindent\textbf{Example.} Consider two actions sets $\{a_{i1}, a_{i2}\}$, $\{a_{j1}, a_{j2}\}$ and time moments $t_i, t_j$. Let $[t_i, t'] = \textit{unsafe}(a_{i1}, a_{j1}) \cap \textit{unsafe}(a_{i1}, a_{j2})$ and $[t_i, t''] = \textit{unsafe}(a_{i2}, a_{j1}) \cap \textit{unsafe}(a_{i2}, a_{j2}])$, where $\textit{unsafe}$ denotes the unsafe interval of $a_{ik}$ starting at $t_i$ w.r.t. $a_{j\ell}$ starting at $t_j$. Then, the MC associated with the first action set is $\{(a_{i1}, [t_i, t')); (a_{i2}, [t_i, t'')) \}$. It dictates that the agent is not allowed to perform either $a_{i1}$ or $a_{i2}$ in the respective time intervals. 
Fig.~\ref{fig:MC-example} shows an illustrative example.

\vspace{3pt}

The problem with MC1 is that in AA-MAPF the number of mutually-conflicting actions may be large. Thus, the intersection of their unsafe intervals may be to short. 
For example, in Fig.~\ref{fig:MC-example} the unsafe interval of action $a_{11}$ w.r.t. $a_{21}$ is $[0; 1.106)$, while its unsafe interval w.r.t. the MC that includes both $a_{21}$ and $a_{22}$ is much shorter --- only $[0; 0.522)$. A shorter unsafe interval means the resulting constraint is weaker. Consequently, more high-level iterations may be required until a solution is found. To mitigate this issue we suggest two modifications. 

\begin{figure}[t]
    \centering
    \includegraphics[width=\linewidth]{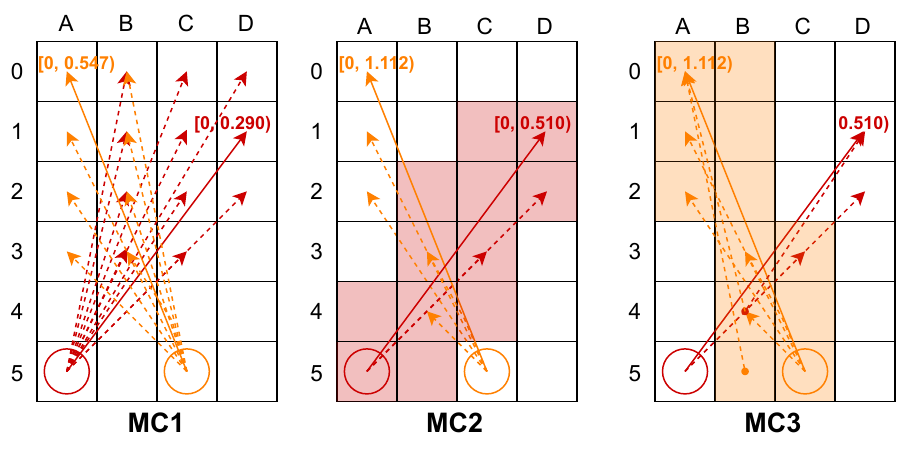}
    \caption{Actions comprising different versions of multi-constraints. Please note that the time intervals of the actions in MC2 and MC3 are notably larger compared to MC1. Thus MC2 and MC3 are expected to exhibit more pruning power.}
    \label{fig:differentMCs}
\end{figure}

First, we suggest to consider only a limited set of actions to form $A_i$ ($A_j$ similarly). The source of each action should coincide with the source of the initially conflicting action $a_i$. The destination must be a vertex that is swept by the agent when executing $a_i$, i.e. the agent's body intersects this vertex (grid cell). Thus the resulting actions form a ``stripe'' along $a_i$ -- see Fig.~\ref{fig:differentMCs} (the cells that form a stripe are highlighted).

The second modification is to filter the actions from the MC that lead to trimming the original unsafe interval. In particular, if for some action $a'_j \in A_j$ it holds that $\textit{unsafe}(a_i, a'_j) \not\subset \textit{unsafe}(a_i, a_j)$ then $a'_j$ is excluded from $A_j$. The combination of those two enhancements is dubbed MC2.

The pseudo-code of constructing MC2 is shown in Alg.~\ref{alg:MC}. After initialization (Lines 1-3), we identify the swept cells (Line 4) and form the candidate sets of actions $\hat{A_i}, \hat{A_j}$ (Line 5). 
Then (Line 6) we merge them into $\hat{A}$ in such a way that the actions from $\hat{A_i}$ and $\hat{A_j}$ alternate (this merging procedure is denoted as $zip$ in code). Next, we iterate over $\hat{A}$ and filter out actions that lead to trimming the unsafe intervals (Lines 10-14).
After pruning we iterate over the actions once again to compute their unsafe intervals (Lines 15-21). The resulting sets of action-interval pairs form MC2.

\begin{algorithm}[t]
    \KwIn{Conflicting actions (and times) $a_i, t_i, a_j, t_j$}
    \KwOut{A pair of multi-constraints $\{A_i, A_j\}$}

    $[t_i, t') \gets$ unsafe interval of $a_i$ w.r.t. $(a_j, t_j)$\;
    $[t_j, t'') \gets$ unsafe interval of $a_j$ w.r.t. $(a_i, t_i)$\;
    $A_i \gets \emptyset$; $A_j \gets \emptyset$\;
    $C_{\{i,j\}} \gets$ vertices swept by $a_{\{i,j\}}$\;
    $\hat{A}_{\{i, j\}} \gets$ valid actions that start at the source of $a_{\{i, j\}}$ and end in $C_{\{i, j\}}$\;
    $\hat{A} \gets zip(\hat{A_i},\hat{A_j})$\;
    
    \For{\upshape\textbf{each} $\hat{a}\in \hat{A}$}
    {
        \eIf{$source(\hat{a}) = source(a_i)$}
        {
            \If{IsConflict($\hat{a}, A_j$)=true}
            {
                $[t_j, \hat{t}) \gets$ unsafe interval of $a_j$ w.r.t. $(\hat{a}, t_i)$\;
                \If{$t'' > \hat{t} $}
                {
                    remove $\hat{a}$ from $\hat{A_i}$\;
                }
            
            }
        }
        {
            handling $a_j$ similarly (Lines 9-12)\;
        }    
    }

    \For{\upshape\textbf{each} $a\in \hat{A_i}$}
    {
        $t_{min} \gets \infty$\;
        \For{\upshape\textbf{each} $a'\in \hat{A_j}$}
        {
            $[t_i, t_{cur}) \gets$ unsafe interval of $a$ w.r.t. $(a', t_j)$\;
            $t_{min} \gets \min\{t_{min}, t_{cur}\}$\;
        }
        $A_i \gets A_i \cup \{(a, [t_i, t_{min}))\}$\;
        
    }

    Process $\hat{A_j}$ similarly (Lines 15-20)\;
    
    \Return $\{A_i, A_j\}$
	
    \caption{Forming MC2 for AA-CCBS} 
\label{alg:MC}
\end{algorithm}

Next, we suggest one more way of composing the action sets of MC, when we include into the constrained set actions that start at the \emph{different} graph vertices. This is possible as by definition the set of actions comprising $A_i$ and $A_j$ must be mutually conflicting but the definition does not specify what source and target vertices of these actions should be. In specifics, we enlarge the action sets of MC2 with the actions that have the same target vertex (as the original conflicting actions) and the source vertices that lie on the same stripe as before -- see Fig.~\ref{fig:differentMCs} (right). The rationale is that these actions are likely to lead to collisions between the same agents in nearly the same place. Thus it is natural to include them in the same multi-constraint. The pseudo-code for constructing MC3 largely repeats the one for MC2 (with certain additions) and is omitted for the sake of space.

\subsection{Disjoint Splitting}

Disjoint splitting (DS) is a powerful technique for reducing search effort in CBS~\cite{li2019disjoint} that was shown to be effective in CCBS as well~\cite{andreychuk2021improving}. In DS for CCBS, a conflict $(a_i, a_j, t_i, t_j)$ is resolved in two ways: one CCBS child gets a regular (negative) constraint on agent $i$, while the other one - a \emph{positive} constraint to agent $i$ and a negative constraint to agent $j$.
The positive constraint comes in the conventional CCBS form of $(a, [t, t'))$ but dictates the agent must perform action $a$ in a time moment belonging to $[t, t')$. 
Thus action $a$ becomes a \textit{landmark} in the sense that the low-level search has to, first, find a plan to the source of $a$, then execute $a$, and then plan to the goal (or to the next landmark). 

In this work, we suggest two ways to implement DS for AA-CCBS: using DS as-is and using DS with MC. The first one is straightforward, while the second requires more attention. 

To integrate DS with MCs that start at the same vertex but end in different ones, like MC1 or MC2, we need to impose a combined positive constraint that will dictate an agent to perform \emph{any} of the actions comprising the constrained set. Thus, we need to reason over the multiple distinct start locations for the consecutive search, which is not trivial. Integrating DS with MC3 is even more involved as here we need to handle landmarks that are composed of the actions that start at different locations.

To simplify the integration of DS and MC in AA-CCBS we suggest to avoid creating landmarks out of multi-constraints, while still operating with conventional negative multi-constraints. In particular, consider a conflicting pair of timed actions, $(a_i, t_i, a_j, t_j)$. As in the original DS, we impose a singular negative constraint on the first agent in the left CCBS child and the corresponding singular positive constraint in the right child. Additionally, the second agent gets a negative multi-constraint in the right child, which is computed w.r.t. the action $(a_i, t_i)$ only. In this way, there is no need to handle positive MCs, while the regular (singular) positive constraint is enforced with the negative MC (in one of the descendants in CCBS search tree).

\begin{theorem}
    AA-CCBS with any of the enhancements -- MC1, MC2, MC3, and DS --  is sound, solution-complete, and optimal, i.e., it is guaranteed to return a valid solution is such exists, and the solution it returns is cost-optimal.  
    \label{the:optimality}    
\end{theorem}
\noindent \textbf{Proof outline:} 
Proving soundness, solution-completeness, and optimality for vanilla AA-CCBS is trivial, following the proofs of CCBS~\cite{andreychuk2022multi} and TO-AA-SIPP~\cite{yakovlev2021towards}. 
To prove that the proposed improvements (MC1, MC2, MC3, and DS) preserve these properties, it is sufficient to show that AA-CCBS with them always resolves conflicts with a \emph{sound pair of constraints}~\cite{atzmon2020robust}. 
 A pair of constraints is sound if in any valid solution, at least one of these constraints is satisfied. 
 That is, any solution in which both constraints are violated is not valid. 
 In MC1, MC2, and MC3, by construction, 
 for every pair of multi-constraints $A_i$ and $A_j$ they return it holds that any pair of constraints $(a_i^*,T_i^*)\in A_i$ and $(a_j^*, T_j^*)\in A_j$ form a sound pair of constraints. 
 Thus, any solution that violates a constraint in $A_i$ and a constraint in $A_j$ cannot be valid. 
A similar argument holds for DS due to the negative constraints it imposes.

\begin{figure*}[tb!]
    \centering
    \begin{subfigure}[b]{0.29\textwidth}
        \centering
        \includegraphics[width=\textwidth]{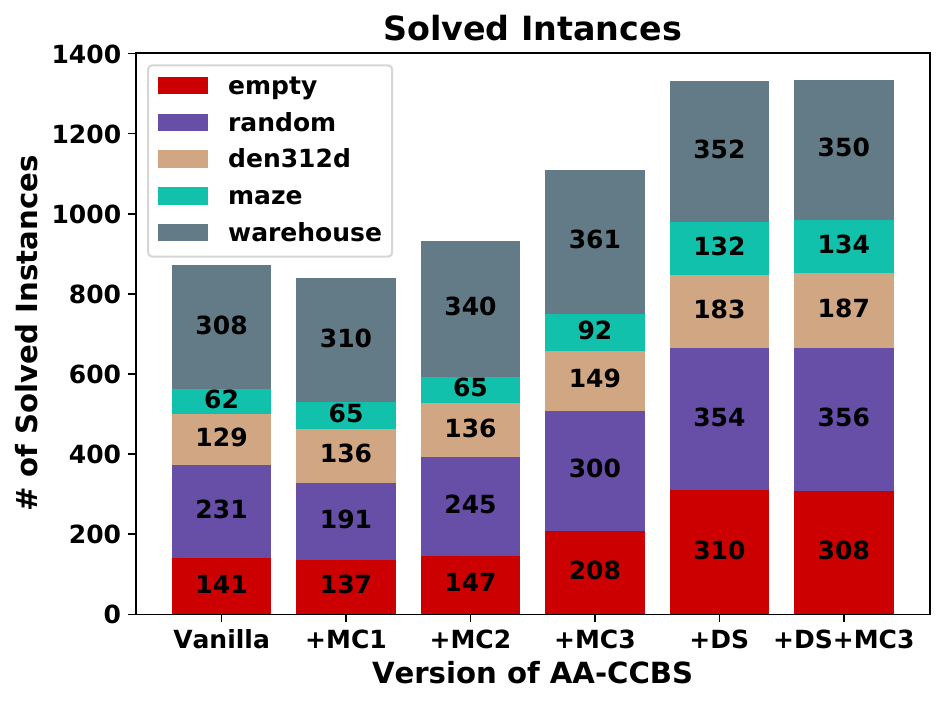}
    \end{subfigure}
    \begin{subfigure}[b]{0.34\textwidth}
        \centering
        \includegraphics[width=\textwidth]{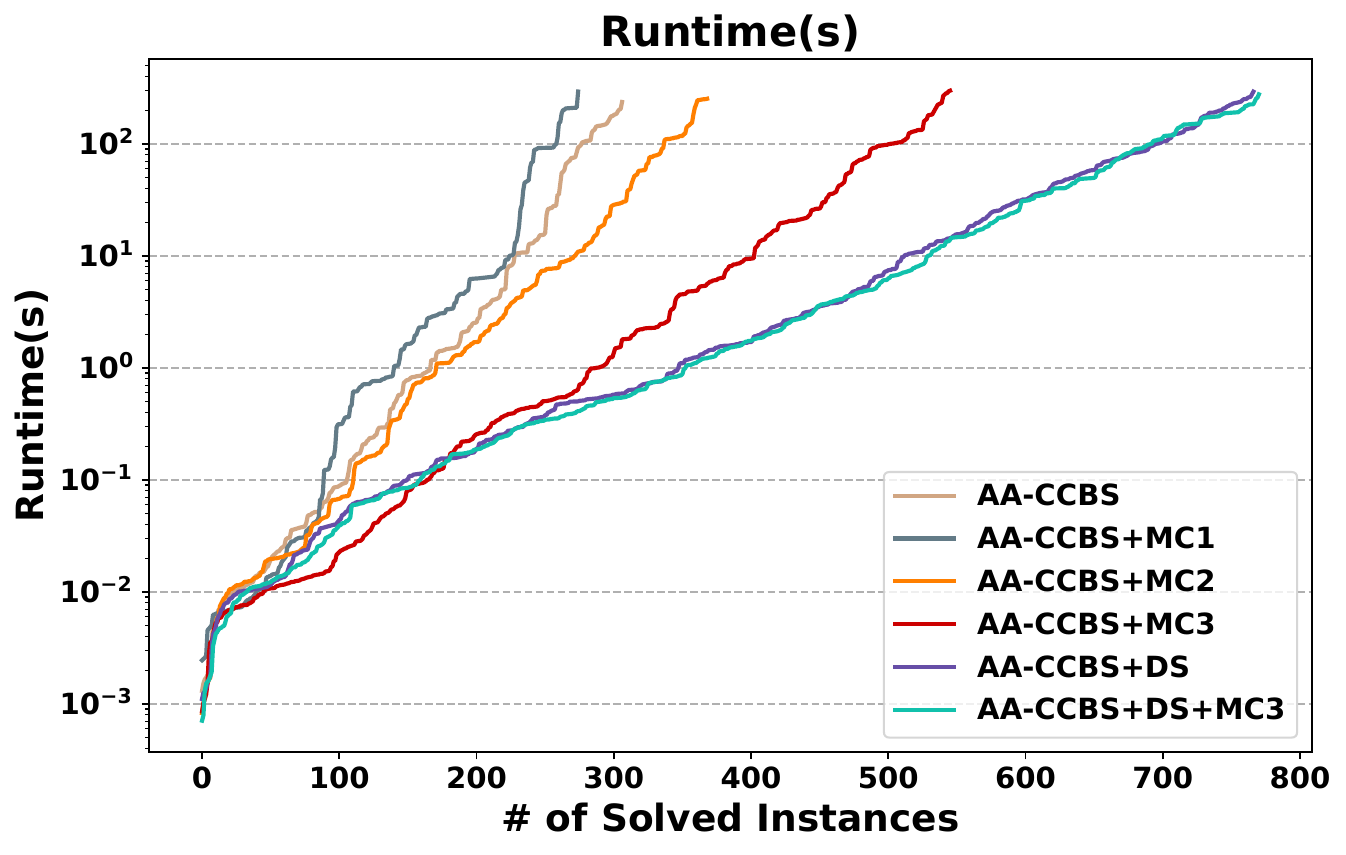}
    \end{subfigure}
    \begin{subfigure}[b]{0.33\textwidth}
        \centering
        \includegraphics[width=\textwidth]{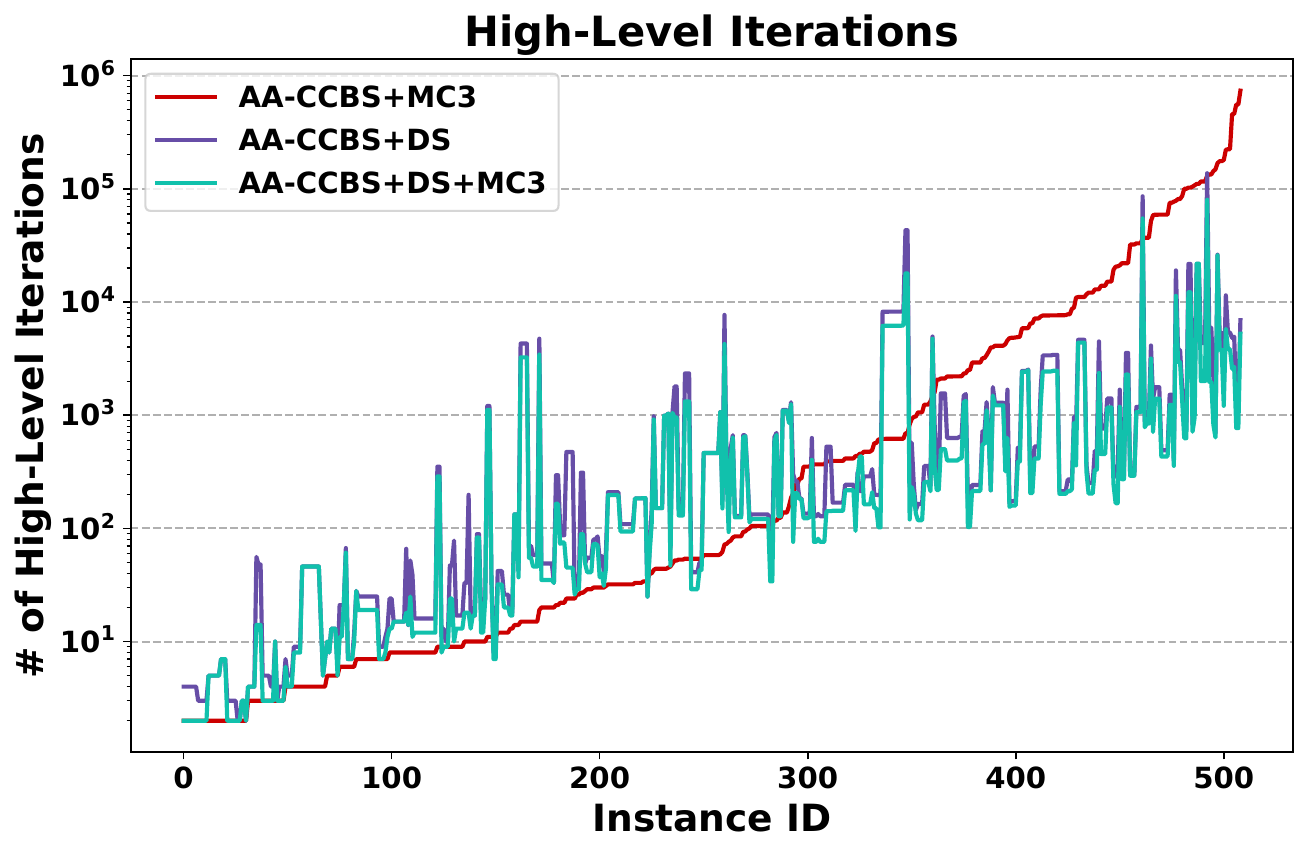}
    \end{subfigure}
    \caption{Evaluation of different variants of AA-CCBS. \textit{Left pane} shows the amount of totally solved instances (within a 5 min. time cap) by each version of AA-CCBS. \textit{Middle pane} shows the time taken to find a solution (Y-axis is in log-scale). Each data point tells how many instances (X-axis) the algorithm is capable to solve within a certain time (Y-axis). \textit{Right pane} demonstrates how many high-level iterations are needed to solve different MAPF problem instances. X-axis shows instance id. Each data point tells how many high-level iterations were made by the algorithm (Y-axis) on a particular problem instance (X-axis).}
    \label{fig:optimal_results}
\end{figure*}

\section{Experimental Results}

We implemented AA-CCBS in C++\footnote{https://github.com/PathPlanning/Continuous-CBS/tree/AA-CCBS} and conducted the evaluation on different maps from the MovingAI benchmark~\cite{stern2019multi}: \texttt{empty-16-16}, \texttt{random-32-32-20}, \texttt{maze-32-32-4}, \texttt{den321d} and \texttt{warehouse-10-20-10-2-2}. For each map, the benchmark provides 25 scenarios. Each scenario is a set of start-goal locations. We take two agents, run the algorithm, and then increment the number of agents until the algorithm is not able to find a solution within a time limit of 300 seconds. 
In the latter case, we proceed to the next scenario. We evaluate six versions of AA-CCBS: AA-CCBS (vanilla), AA-CCBS+MC1, AA-CCBS+MC2, AA-CCBS+MC3, AA-CCBS+DS, AA-CCBS+DS+MC3. 

Fig.~\ref{fig:optimal_results} shows the results.
The left pane shows the \emph{coverage}, which is the number of solved instances within our 300 seconds time limit. The colors show the breakdowns for each map. 
The results indicate that MC1 and MC2 do not provide a significant improvement over vanilla AA-CCBS, while
MC3, and especially DS and DS+MC3, outperform it significantly. 
The central pane provides a finer-grained analysis, plotting the number of instances solved ($x$-axis) for different time limits ($y$-axis). We filtered out trivial instances where 
the initial any-angle paths of the agents do not contain any conflict. 
This analysis shows that MC3 outperforms all other variants of AA-CCBS if the time limit is between 0.01 and 0.1 seconds, while DS and DS+MC3 outperform MC3 when the time limit is higher. 
This suggests MC3 is better suited for quickly solving easier instances while DS is better for harder instances. 
To get a better understanding of this behavior, the right pane of Fig.~\ref{fig:optimal_results} depicts the number of the high-level iterations of AA-CCBS with MC3, DS, DS+MC3 for each MAPF instance that was solved by all these methods. The instances are sorted by the number of iterations of MC3 (so its line monotonically increases) and the $x$-axis corresponds to distinct instances. 
As can be seen, in nearly half of the instances MC3 requires fewer high-level iterations than DS and DS+MC3. These are the cases that in general required fewer high-level iterations. 
Overall, our results confirm that both MC and DS are valuable techniques to enhance AA-CCBS performance. MC3 is well-suited for the easy instances while DS is better for the harder ones. 

A comparison of AA-CCBS with its non any-angle variant is presented in Appendix.

\begin{figure}[t!]
    \centering
    \includegraphics[width=1\linewidth]{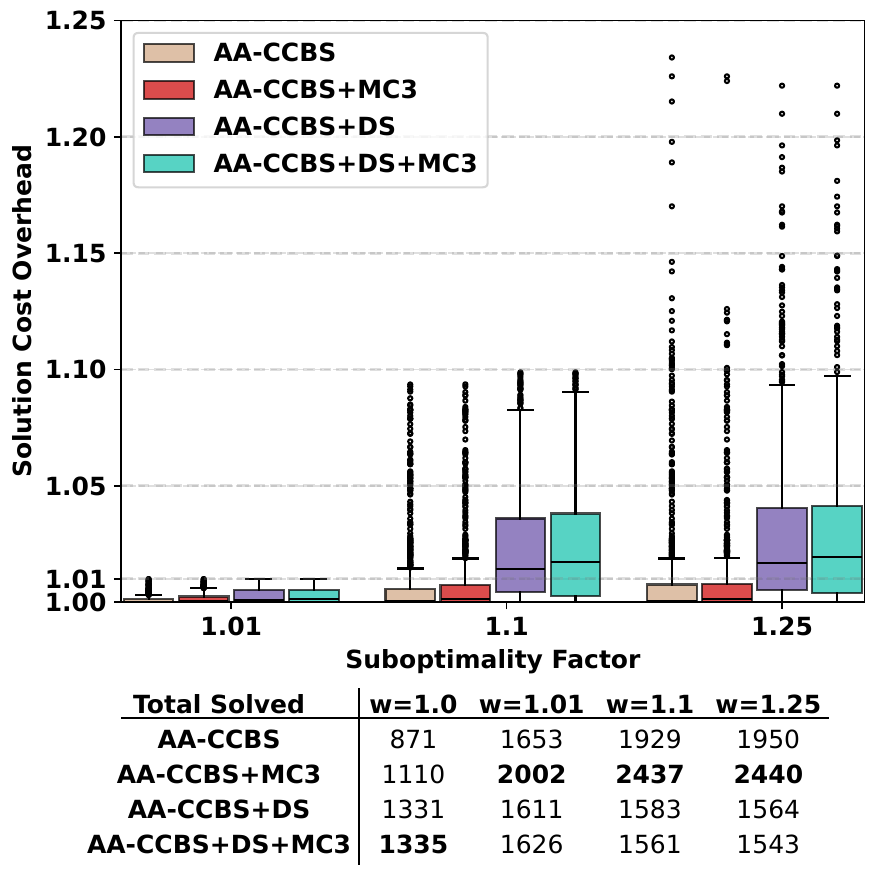}
    \caption{The results of the evaluation of AA-CCBS and its modifications with different suboptimality factors.
    }
    \label{fig:suboptimal_results}
\end{figure}

\section{Bounded-Suboptimal AA-CCBS}
Given a real number $w>1$, a bounded suboptimal (BS) solution is a solution whose cost does not exceed the cost of an optimal solution by more than a factor of $w$. 
Finding a BS solution is often easier than finding an optimal one. 
Creating a BS variant of AA-CCBS with any of our enhancements can be done by replacing the best-first search in the high-level of AA-CCBS with a \emph{Focal Search}~\cite{pearl1982studies}.\footnote{Creating a BS low-level search is not trivial in is beyond the scope of the paper.} In more detail, in each high-level iteration we create a list of nodes called the \emph{FOCAL}. 
FOCAL contains all the generated nodes whose cost does not exceed $w\cdot f_{min}$, where $f_{min}$ 
is the cost of the node regular AA-CCBS would have expanded. 
Our BS version of AA-CCBS chooses to expand the node in FOCAL with the smallest number of conflicts, breaking ties in favor of nodes with more constraints. This forces the search to resolve conflicts more greedily, while maintaining the required suboptimality bound. 

We implemented this BS version of AA-CCBS with the three best-performing enhancements, DS, MC3, DS+MC3, and evaluated them on the same problem instances as before. 
Fig.~\ref{fig:suboptimal_results} shows experimental results over the same problem instances as above across all maps using our BS version of AA-CCBS for suboptimality bounds $w=1.01$, $1.1$, and $1.25$
with the three best-performing enhancements, DS, MC3, DS+MC3. 
Fig.~\ref{fig:suboptimal_results}(top) depicts a box-and-whiskers plot of the solution cost overhead w.r.t. the SOC of the optimal solution. 
Fig.~\ref{fig:suboptimal_results}(bottom) shows the number of solved instances within our 300 seconds time limit (coverage). 

In terms of coverage, MC3 notably increases the performance of AA-CCBS when searching for bounded suboptimal solutions (i.e., when $w > 1$), while DS variants are not even able to outperform vanilla AA-CCBS. The poor results of DS may be due to the positive constraints it imposes, which often do not lead to immediately eliminating the conflict that causes these constraints. When searching for optimal solutions this is compensated by reducing the branching factor. However, the search for suboptimal solutions is more greedy by design and the branching factor becomes much less important. Thus, DS loses much of its power, while retaining the drawback of not immediately eliminating the conflict. At the same time, MC3 imposes constraints on a larger number of actions, and as the search greedily goes deeper it actually eliminates the conflict. 

Moreover, due to the requirement to satisfy all the imposed positive constraints, the actual costs of the solutions found by DS versions are significantly higher than the ones of MC3. 
This can be seen in Fig.~\ref{fig:suboptimal_results}(top). 
The actual overhead in solution costs of MC3 version is not higher than 2\% on average even when the suboptimality bound is $w=1.25$. 
The obtained results provide further support that MC3 is a useful technique: not only does it speed the search for optimal solutions on easy instances, but it also significantly increases the performance of bounded-suboptimal AA-CCBS.

\section{CONCLUSION AND FUTURE WORK}
In this work, we presented AA-CCBS, the first optimal any-angle MAPF algorithm. We showed how to incorporate existing CCBS enhancements, namely Disjoint Splitting and Multi-Constraints, into AA-CCBS to allow it to scale better. Specifically, we introduced three ways to implement Multi-Constraints, providing stronger pruning of the high-level search tree. AA-CCBS can also be easily adapted to return bounded-suboptimal solutions. 
Experimental results confirm that the suggested enhancements have a huge impact on the algorithm's performance. 
Future work can explore more sophisticated procedures of forming multi-constraints, identifying which sets of actions should be considered in each step, as well as adapting incremental search techniques in the any-angle low-level search.

\section*{ACKNOWLEDGMENT}
This research is partially funded by ISF grant \#1238/23 to Roni Stern and by the Ministry of Science and Higher Education of the Russian Federation (Project No. 075-15-2024-544).


\bibliographystyle{ieeetr} 
\bibliography{references}

\begin{thebibliography}{10}

\bibitem{wurman2008coordinating}
P.~R. Wurman, R.~D'Andrea, and M.~Mountz, ``Coordinating hundreds of
  cooperative, autonomous vehicles in warehouses,'' {\em AI magazine}, vol.~29,
  no.~1, pp.~9--9, 2008.

\bibitem{parks2022intersection}
A.~Parks-Young and G.~Sharon, ``Intersection management protocol for mixed
  autonomous and human-operated vehicles,'' {\em IEEE Trans Intell Transp
  Syst}, vol.~23, no.~10, pp.~18315--18325, 2022.

\bibitem{ma2017feasibility}
H.~Ma, J.~Yang, L.~Cohen, T.~K.~S. Kumar, and S.~Koenig, ``Feasibility study:
  Moving non-homogeneous teams in congested video game environments,'' in {\em
  AIIDE 2017}, pp.~270--272, 2017.

\bibitem{yu2016optimal}
J.~Yu and S.~M. LaValle, ``Optimal multirobot path planning on graphs: Complete
  algorithms and effective heuristics,'' {\em IEEE Trans Robot}, vol.~32,
  no.~5, pp.~1163--1177, 2016.

\bibitem{sharon2015cbs}
G.~Sharon, R.~Stern, A.~Felner, and N.~R. Sturtevant, ``Conflict-based search
  for optimal multi-agent pathfinding,'' {\em Artif Intell}, 2015.

\bibitem{surynek2016efficient}
P.~Surynek, A.~Felner, R.~Stern, and E.~Boyarski, ``Efficient sat approach to
  multi-agent path finding under the sum of costs objective,'' in {\em ECAI
  2016}, pp.~810--818, 2016.

\bibitem{lam2022bcp}
E.~Lam, P.~{Le Bodic}, D.~Harabor, and P.~J. Stuckey,
  ``Branch-and-cut-and-price for multi-agent path finding,'' {\em Comput Oper
  Res}, vol.~144, p.~105809, 2022.

\bibitem{stern2019multi}
R.~Stern, N.~R. Sturtevant, A.~Felner, S.~Koenig, H.~Ma, T.~T. Walker, J.~Li,
  D.~Atzmon, L.~Cohen, T.~S. Kumar, {\em et~al.}, ``Multi-agent pathfinding:
  Definitions, variants, and benchmarks,'' in {\em SoCS 2019}, pp.~151--158,
  2019.

\bibitem{andreychuk2022multi}
A.~Andreychuk, K.~Yakovlev, P.~Surynek, D.~Atzmon, and R.~Stern, ``Multi-agent
  pathfinding with continuous time,'' {\em Artif Intell}, vol.~305, p.~103662,
  2022.

\bibitem{nash2007}
A.~Nash, K.~Daniel, S.~Koenig, and A.~Felner, ``Theta*: Any-angle path planning
  on grids,'' in {\em AAAI 2007}, pp.~1177--1183, 2007.

\bibitem{harabor2013optimal}
D.~Harabor and A.~Grastien, ``An optimal any-angle pathfinding algorithm,'' in
  {\em ICAPS 2013}, pp.~308--311, 2013.

\bibitem{yakovlev2021towards}
K.~Yakovlev and A.~Andreychuk, ``Towards time-optimal any-angle path planning
  with dynamic obstacles,'' in {\em ICAPS 2021}, pp.~405--414, 2021.

\bibitem{yakovlev2017any}
K.~Yakovlev and A.~Andreychuk, ``Any-angle pathfinding for multiple agents
  based on sipp algorithm,'' in {\em ICAPS 2017}, pp.~586--594, 2017.

\bibitem{li2019disjoint}
J.~Li, D.~Harabor, P.~J. Stuckey, A.~Felner, H.~Ma, and S.~Koenig, ``Disjoint
  splitting for multi-agent path finding with conflict-based search,'' in {\em
  ICAPS 2019}, pp.~279--283, 2019.

\bibitem{walker2020generalized}
T.~T. Walker, N.~R. Sturtevant, and A.~Felner, ``Generalized and sub-optimal
  bipartite constraints for conflict-based search,'' in {\em AAAI 2020},
  pp.~7277--7284, 2020.

\bibitem{yakovlev2019prioritized}
K.~Yakovlev, A.~Andreychuk, and V.~Vorobyev, ``Prioritized multi-agent path
  finding for differential drive robots,'' in {\em ECMR 2019}, pp.~1--6, 2019.

\bibitem{ma2019lifelong}
H.~Ma, W.~H\"{o}nig, T.~K.~S. Kumar, N.~Ayanian, and S.~Koenig, ``Lifelong path
  planning with kinematic constraints for multi-agent pickup and delivery,'' in
  {\em AAAI 2019}, pp.~7651--7658, 2019.

\bibitem{honig2016multi}
W.~H{\"o}nig, T.~S. Kumar, L.~Cohen, H.~Ma, H.~Xu, N.~Ayanian, and S.~Koenig,
  ``Multi-agent path finding with kinematic constraints.,'' in {\em ICAPS
  2016}, pp.~477--485, 2016.

\bibitem{ali2021prioritized}
Z.~A. Ali and K.~Yakovlev, ``Prioritized sipp for multi-agent path finding with
  kinematic constraints,'' in {\em ICR 2021}, pp.~1--13, 2021.

\bibitem{YanRAL24}
J.~Yan and J.~Li, ``Multi-agent motion planning with bezier curve optimization
  under kinodynamic constraints,'' {\em IEEE Robot Autom Lett}, vol.~9, no.~3,
  pp.~3021--3028, 2024.

\bibitem{ma2019searching}
H.~Ma, D.~Harabor, P.~J. Stuckey, J.~Li, and S.~Koenig, ``Searching with
  consistent prioritization for multi-agent path finding,'' in {\em AAAI
  2019)}, pp.~7643--7650, 2019.

\bibitem{solis2021representation}
I.~Solis, J.~Motes, R.~Sandstr{\"o}m, and N.~M. Amato, ``Representation-optimal
  multi-robot motion planning using conflict-based search,'' {\em IEEE Robot
  Autom Lett}, vol.~6, no.~3, pp.~4608--4615, 2021.

\bibitem{debord2018trajectory}
M.~Debord, W.~H{\"o}nig, and N.~Ayanian, ``Trajectory planning for
  heterogeneous robot teams,'' in {\em IROS 2018}, pp.~7924--7931, 2018.

\bibitem{honig2018trajectory}
W.~H{\"o}nig, J.~A. Preiss, T.~S. Kumar, G.~S. Sukhatme, and N.~Ayanian,
  ``Trajectory planning for quadrotor swarms,'' {\em IEEE Trans Robot},
  vol.~34, no.~4, pp.~856--869, 2018.

\bibitem{kottinger2022conflict}
J.~Kottinger, S.~Almagor, and M.~Lahijanian, ``Conflict-based search for
  multi-robot motion planning with kinodynamic constraints,'' in {\em IROS
  2022}, pp.~13494--13499, 2022.

\bibitem{moldagalieva2023db}
A.~Moldagalieva, J.~Ortiz-Haro, M.~Toussaint, and W.~H{\"o}nig, ``{db-CBS}:
  Discontinuity-bounded conflict-based search for multi-robot kinodynamic
  motion planning,'' in {\em ICRA 2024}, pp.~14569--14575, 2024.

\bibitem{wen2022cl}
L.~Wen, Y.~Liu, and H.~Li, ``Cl-mapf: Multi-agent path finding for car-like
  robots with kinematic and spatiotemporal constraints,'' {\em Rob Auton Syst},
  vol.~150, p.~103997, 2022.

\bibitem{walker2019collision}
T.~T. Walker and N.~R. Sturtevant, ``Collision detection for agents in
  multi-agent pathfinding,'' {\em arXiv preprint arXiv:1908.09707}, 2019.

\bibitem{phillips2011sipp}
M.~Phillips and M.~Likhachev, ``{SIPP}: Safe interval path planning for dynamic
  environments,'' in {\em ICRA 2011}, pp.~5628--5635, 2011.

\bibitem{li2019multi}
J.~Li, P.~Surynek, A.~Felner, H.~Ma, and S.~Koenig, ``Multi-agent path finding
  for large agents,'' in {\em AAAI 2019}, pp.~7627--7634, 2019.

\bibitem{andreychuk2021improving}
A.~Andreychuk, K.~Yakovlev, E.~Boyarski, and R.~Stern, ``Improving
  continuous-time conflict based search,'' in {\em AAAI 2021},
  pp.~11220--11227, 2021.

\bibitem{atzmon2020robust}
D.~Atzmon, R.~Stern, A.~Felner, G.~Wagner, R.~Bart{\'a}k, and N.-F. Zhou,
  ``Robust multi-agent path finding and executing,'' {\em J Artif Intell Res},
  vol.~67, pp.~549--579, 2020.

\bibitem{pearl1982studies}
J.~Pearl and J.~H. Kim, ``Studies in semi-admissible heuristics,'' {\em IEEE
  Trans Pattern Anal Mach Intell}, no.~4, pp.~392--399, 1982.

\end{thebibliography}

\addtolength{\textheight}{-6cm}   

\newpage

\section*{APPENDIX}

\subsection{Additional Experiments}

To better understand what is the effect of allowing any-angle moves for the agents in MAPF we have compared AA-CCBS with its non-any-angle counterpart, i.e. with the original CCBS solver that allowed moves into 4 cardinal directions on the grid.

For this experiment we have used the most advanced version of CCBS described in~\cite{andreychuk2021improving}. It leverages such enhancements as prioritizing conflicts, disjoint-splitting and high-level heuristics. We denote this version as CCBS (cardinal-only). We compare it with the two variants of the proposed any-angle MAPF solver, i.e. AA-CCBS+DS+MC3 and AA-CCBS+MC3 with the suboptimality factor $w=1.25$. These solvers are denoted as AA-CCBS (optimal) and AA-CCBS (suboptimal).

\begin{figure}
    \centering
    \begin{subfigure}[b]{0.23\textwidth}
        \centering
        \includegraphics[width=\textwidth]{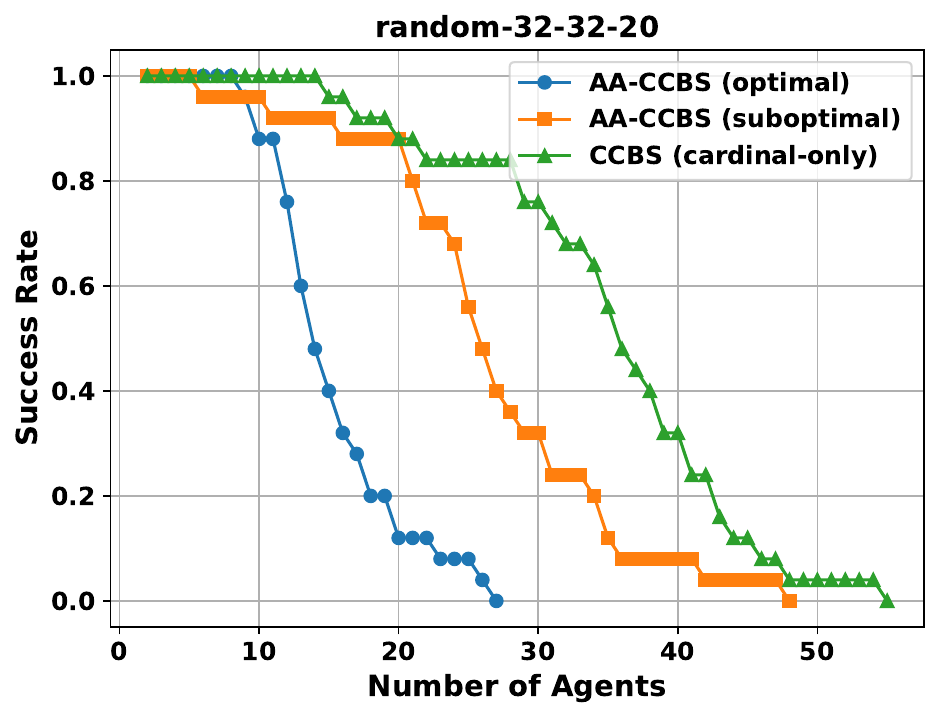}
    \end{subfigure}
    \begin{subfigure}[b]{0.23\textwidth}
        \centering
        \includegraphics[width=\textwidth]{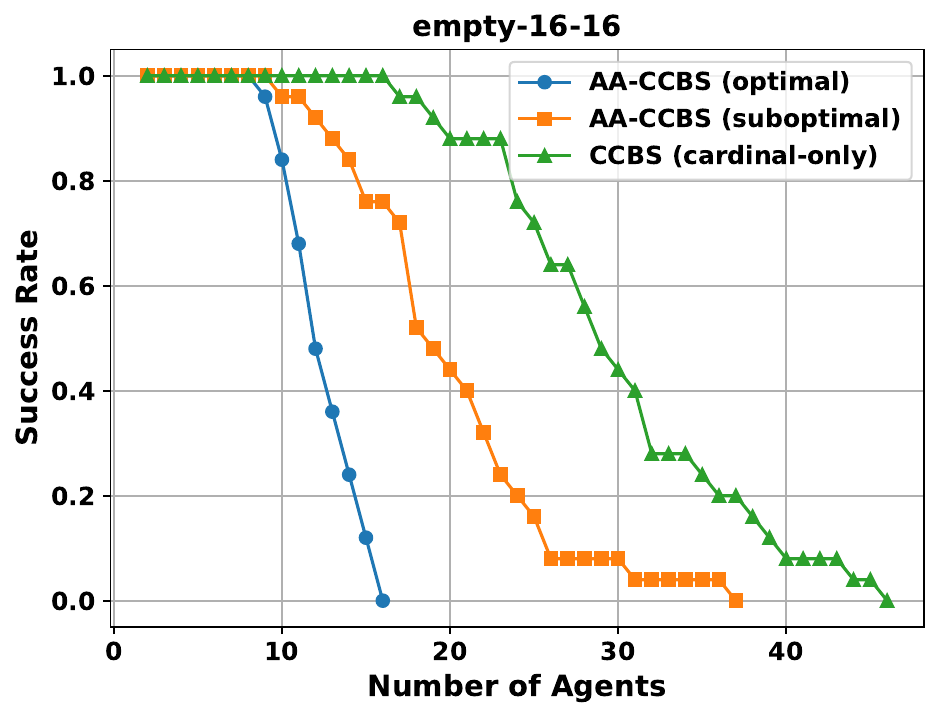}
    \end{subfigure}
    \begin{subfigure}[b]{0.23\textwidth}
        \centering
        \includegraphics[width=\textwidth]{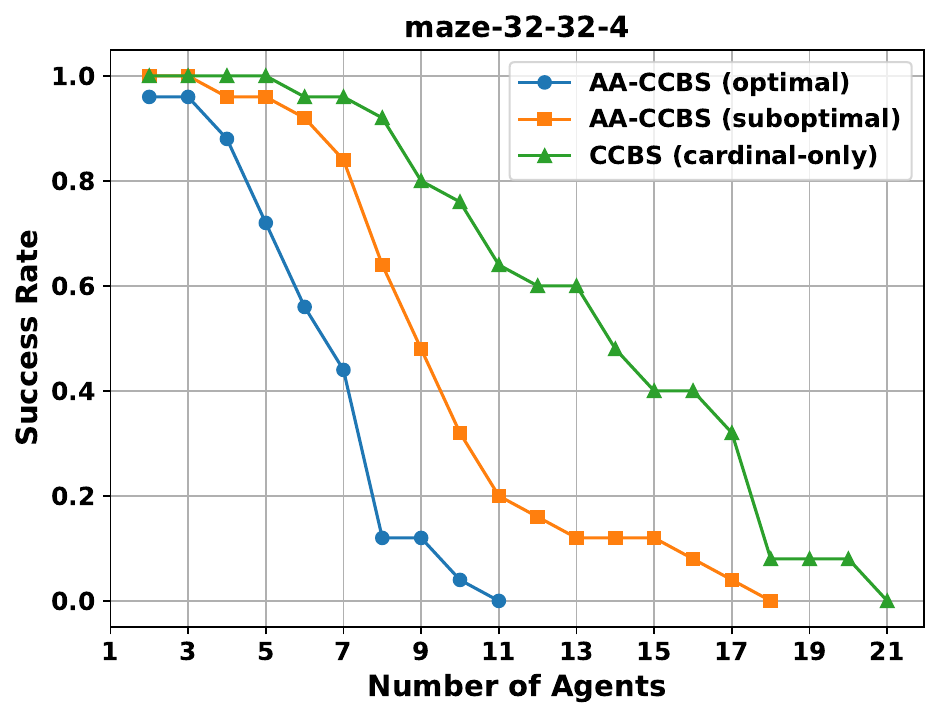}
    \end{subfigure}
    \begin{subfigure}[b]{0.23\textwidth}
        \centering
        \includegraphics[width=\textwidth]{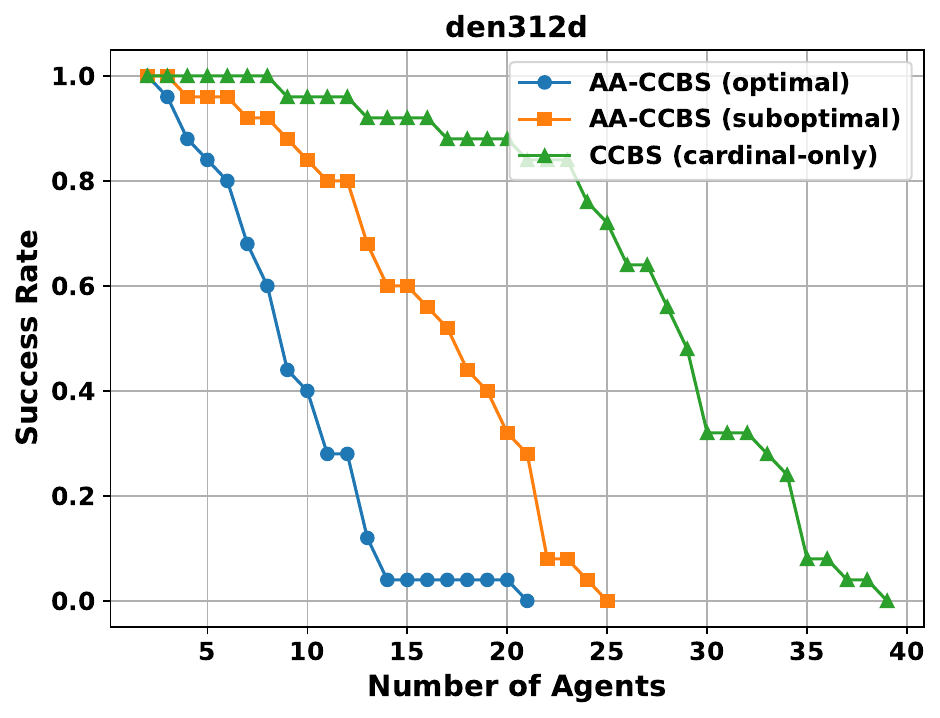}
    \end{subfigure}
    \begin{subfigure}[b]{0.23\textwidth}
        \centering
        \includegraphics[width=\textwidth]{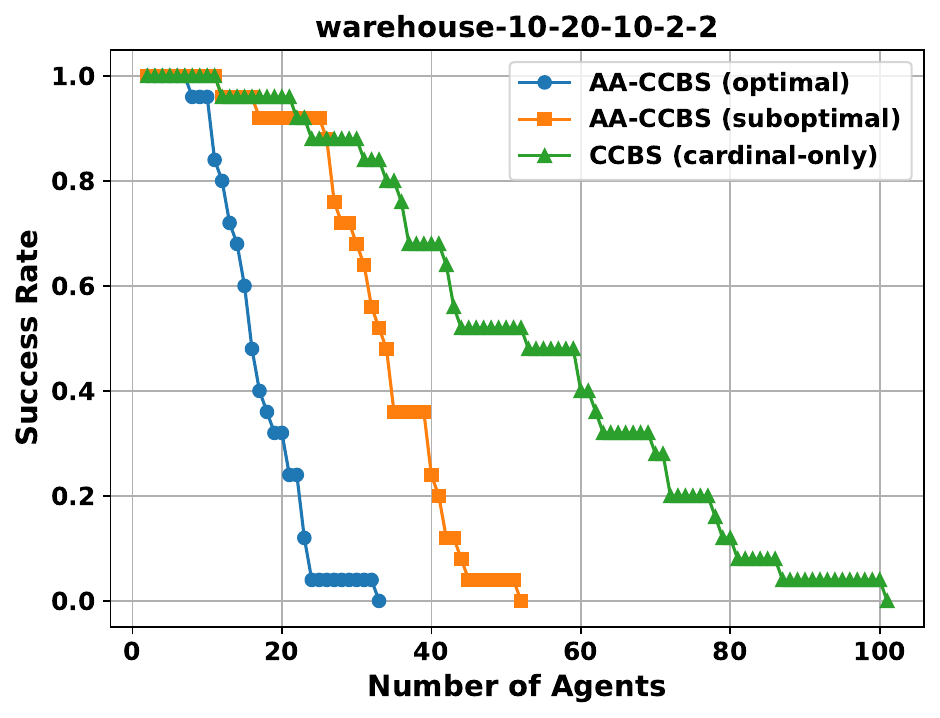}
    \end{subfigure}
    \caption{Success rates per number of agents for AA-CCBS and CCBS on different maps.}
    \label{fig:success_rates}
\end{figure}

The evaluation protocol and setup was the same as in the experiments reported in the main body of the paper (we used the same maps, problem instances, time cap etc.). The results are reported in Table~\ref{tab:ccbs_vc_aaccbs} and Fig.~\ref{fig:success_rates}.

\begin{table}[h!]
\centering
\begin{tabular}{l|ccc}
& CCBS & AA-CCBS & AA-CCBS \\
& (cardinal-only) & (optimal) & (suboptimal) \\
\hline
empty-16-16             & 127.4\% & 100.0\% & 101.4\% \\
random-32-32-20         & 117.6\% & 100.0\% & 101.1\% \\
maze-32-32-4            & 115.6\% & 100.0\% & 100.4\% \\
den312d                 & 117.9\% & 100.0\% & 100.2\% \\
warehouse-10-20-10-2-2  & 111.8\% & 100.0\% & 100.2\% \\
\end{tabular}
\caption{The solution cost of CCBS (cardinal-only) compared to AA-CCBS (optimal) and AA-CCBS (suboptimal). The cost of AA-CCBS (optimal) is taken for 100\%.}
\label{tab:ccbs_vc_aaccbs}
\end{table}

Table~\ref{tab:ccbs_vc_aaccbs} shows the average solution cost relative to AA-CCBS (optimal). Recall that CCBS is an optimal solver so its solutions (under the given movement assumption) can not be improved. Still, as one can see from the table, the gap in solution cost between AA-CCBS (optimal) and CCBS (cardinal-only) can reach $27.4\%$. The lowest difference in solution cost is observed on the warehouse map. This can be explained by the structure of this map that consists of numerous narrow corridors that limit the ability to perform any-angle moves. When the map contains empty areas the gap is getting bigger.

Figure~\ref{fig:success_rates} demonstrates the success rate of the evaluated solvers across all maps. X-axis of each plot shows the number of agents and Y-axis shows the percentage of the problem instances (with this number of agents) successfully solved by a solver within a given time limit (of 300 seconds). 

As CCBS (cardinal only) does not consider any-angle moves its search space is much more compact. Thus it manages to consistently outperform both optimal and suboptimal variants of AA-CCBS in terms of success rate. It is worth noting that AA-CCBS (suboprimal) performs notably better than AA-CCBS (optimal) and, generally, can be considered to provide a good trade off between the solution cost and planning time.

\end{document}